\documentclass[A4paper, 11 pt, conference]{ieeeconf}  
\IEEEoverridecommandlockouts       
\usepackage{geometry}
\usepackage{url}
\usepackage{graphicx}
\usepackage{booktabs}
\usepackage{romannum}
\usepackage{amsmath}
\usepackage{amsfonts}
\usepackage{multirow}
\usepackage[table,xcdraw]{xcolor}
\usepackage{array}
\usepackage{cleveref}
\usepackage{float}
\usepackage[export]{adjustbox} 
\usepackage{algorithm}
\usepackage{balance}
\usepackage{cite}
\usepackage{algpseudocode} 
\usepackage{longtable}
\usepackage{booktabs}
\usepackage[german]{babel}
\usepackage[utf8]{inputenc}
\newcommand{\squeezeup}{\vspace{-2.5mm}}
\UseRawInputEncoding

\begin{document}

\title{\LARGE \bf
Maximizing the Use of Environmental Constraints: \\A Pushing-Based Hybrid Position/Force Assembly Skill \\for Contact-Rich Tasks}



\author{Yunlei Shi$^{1,2}$,  Zhaopeng Chen$^{2,1}$,  Lin Cong$^{1}$, Yansong Wu$^{3}$, Martin Craiu-M\"uller$^{2}$,\\ Chengjie Yuan$^{3}$, Chunyang Chang$^{2}$, Lei Zhang$^{1,2}$, Jianwei Zhang$^{1}$
\thanks{*This research has received funding from the German Research Foundation (DFG) and the National Science Foundation of China (NSFC) in project Crossmodal Learning, DFG TRR-169/NSFC 61621136008, partially supported by European projects H2020 STEP2DYNA (691154) and ULTRACEPT (778602).}
\thanks{{$^{1}$TAMS (Technical Aspects of Multimodal Systems), Department of
Informatics, Universit\"at Hamburg}, {$^{2}$Agile Robots AG}, {$^{3}$Technische Universit\"at M\"unchen.}}}

\maketitle 
\thispagestyle{empty}

\begin{abstract}
The need for contact-rich tasks is rapidly growing in modern manufacturing settings. However, few traditional robotic assembly skills consider environmental constraints during task execution, and most of them use these constraints as termination conditions. In this study, we present a pushing-based hybrid position/force assembly skill that can maximize environmental constraints during task execution. To the best of our knowledge, this is the first work that considers using pushing actions during the execution of the assembly tasks. We have proved that our skill can maximize the utilization of environmental constraints using mobile manipulator system assembly task experiments, and achieve a 100\% success rate in the executions. 
\end{abstract}

\section{INTRODUCTION}
In most advanced computer, communication, and consumer electronics (3C) product manufacturing factories, the vast majority of production processes requiring high-precision operations have been efficiently realized by fixed, dedicated automation systems or position-controlled robots \cite{ljasenko2020dynamic}. The requirement for the quick setup and reprogram of new robotic assembly lines for new products in 3C factories is rapidly growing \cite{kramberger2016learning}, \cite{li2019survey}, \cite{sloth2020towards}. However, dedicated automation systems and position-controlled robots require tedious programming and tuning, thereby prolonging the time to install a new production line. Moreover, they cannot adapt to any unexpected variations in assembly processes \cite{zhu2018robot}.

Collaborative robots (cobots) assist in tackling the challenges posed by onerous setup, programming tasks \cite{safeea2017end}, 
\begin{figure}[htbp]
\centerline{\includegraphics[width=9.0cm,height=5.9cm]{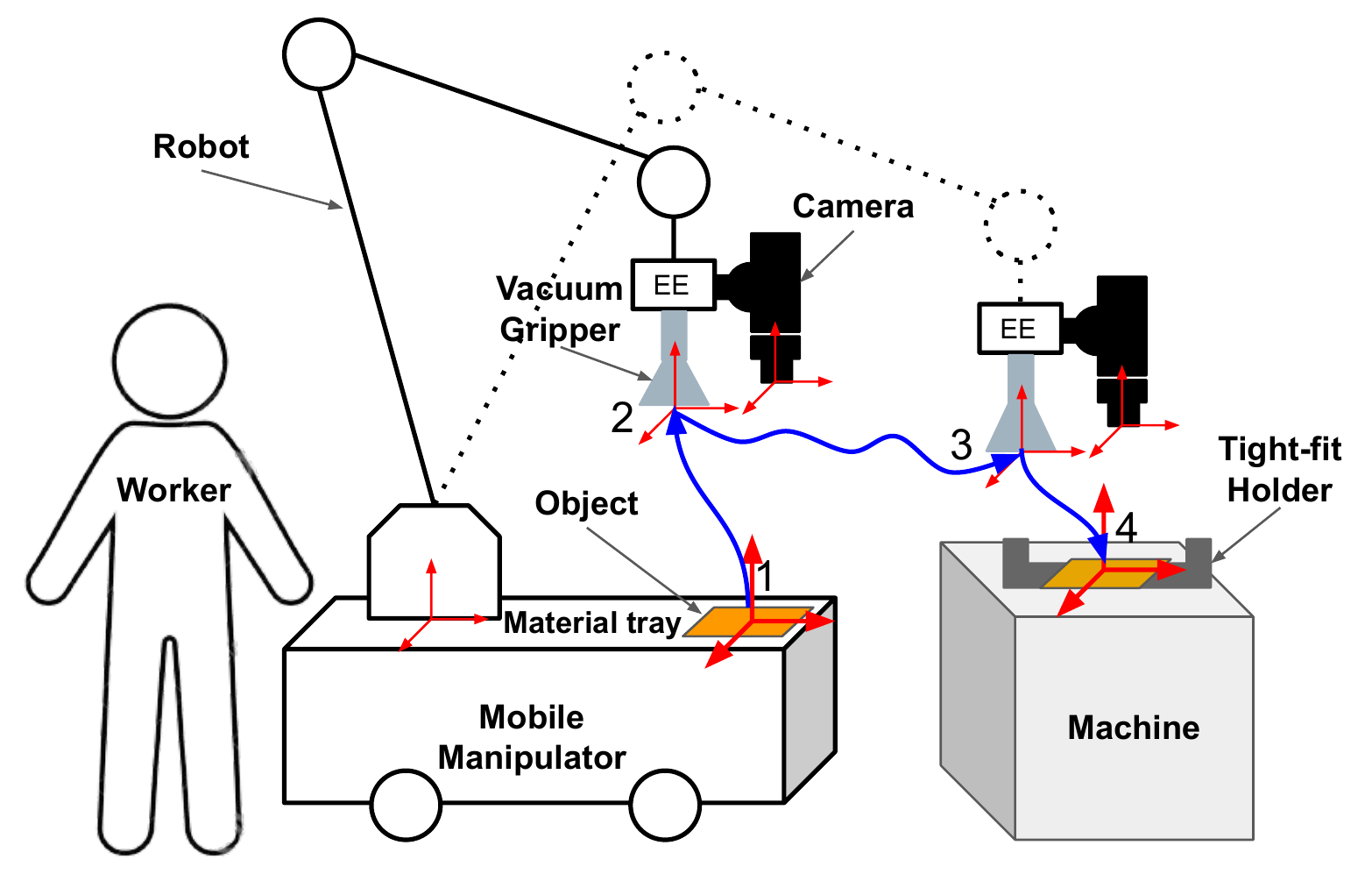}}
\caption{A mobile manipulator system performing a contact-rich tending task using a vacuum gripper.}
\label{fig:system setup}
\squeezeup
\squeezeup
\end{figure}
and unexpected variations in assembly processes through force-control techniques \cite{albu2007dlr}. Mobile manipulator systems were designed to accelerate the long installation and setup times of cobots, as they integrated all the required devices on a ``plug-and-play'' mobile platform \cite{hvilshoj2009mobile}, \cite{wurll2018production}. However, the repeatability of these mobile robots is much worse than that of cobots \cite{su2018positioning}, \cite{li2019survey}.

In this paper, we analyze the constraint case when the object and the environment have relative motion, and thus propose a pushing-based hybrid position/force assembly skill accordingly to address the aforementioned challenge. We demonstrated that the proposed skill can maximize the utilization of environmental constraints through a comparative experiment (\Cref{fig:system setup}).

%
\section{RELATED WORK}\label{RELATED WORK}

\subsection{Robotic Assembly in 3C Manufacturing Industries}
The 3C manufacturing industry is one of the most important industries around the world \cite{zhang2017challenges}. A typical manufacturing process can be divided into four main phases: module manufacturing, assembly, testing and packaging, where the latter \begin{figure}[htbp]
\centerline{\includegraphics[width=8.6cm,height=5.9cm]{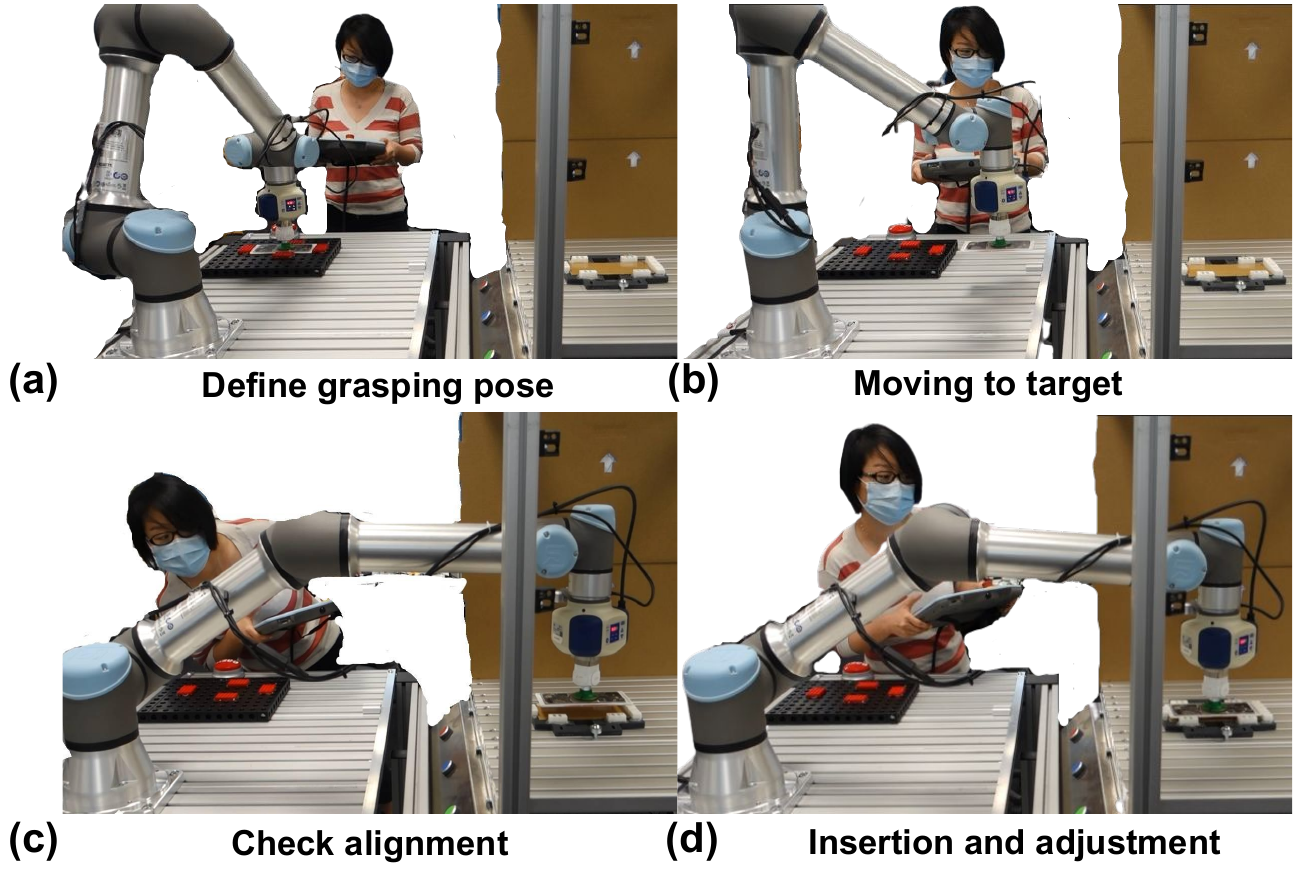}}
\caption{Teaching process with a teach pendant. The users need to adjust the pose of the object using visual feedback, thus causing a contact force between the object and the environment}
\label{fig:teach process}
\squeezeup
\end{figure}three majorly depend dependent on human labor \footnote{https://ri.ust.hk/3C-Industry-Assembly}. Robots have been extensively used in structured production environments to ensure consistent yields and efficient manufacturing \cite{fang2016dual}. However, reinstalling, setting up, and reprogramming robots take several weeks because they cannot rapidly adapt and reorganize with changes in assembly lines.


\subsection{Mobile Manipulator Programming}\label{Fast and easy Programming with mobile manipulator}
Mobile manipulators were designed to accelerate the setup of robots by integrating the required task devices into a mobile platform. Additionally, several mobile manipulators have been developed and deployed in real manufacturing environments. Traditional robot programming is complex. For example, a typical ``pick-and-place'' teaching process comprises several steps (\Cref{fig:teach process}) \cite{shi2021combining}.


Skill-based task programming methods were designed to enable an easy set up for mobile manipulators \cite{pedersen2016robot}, \cite{brunner2016rafcon}, in which the predefined functional blocks are adapted to a specific task via several parameters  \cite{bogh2012does}. Although numerous skills have been developed to simplify mobile manipulator programming \cite{pedersen2016robot}, the research and development of the contact-rich task skills have not been adequately reported.
\begin{figure}[htbp]
\centerline{\includegraphics[width=8.6cm,height=3.5cm]{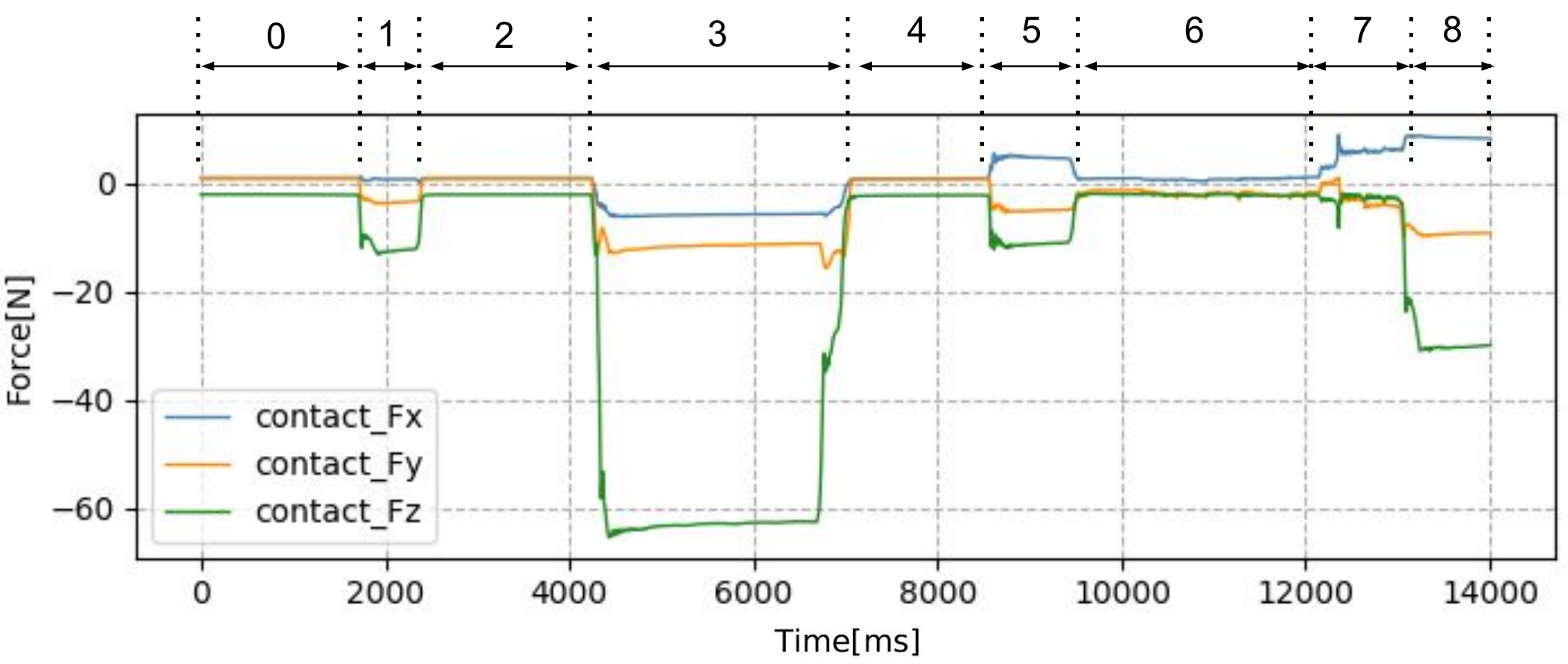}}
\caption{Contact force curve of the object with the holder: approaching, adjusting, and inserting. In phases 0, 2, 4, and 6, the object is not in contact with the holder. In phases 1, 3, 5, and 7, the object is in contact with the holder but not totally inside it. In phase 8, the object is in contact with the holder and totally inside it, and the residual contact forces in \textbf{X,} \textbf{Y,} and \textbf{Z} directions.}
\label{fig:contact force curve}
\squeezeup
\end{figure}
\subsection{Operational Space Position/Force Control}
Direct/indirect force control techniques have been previously used to overcome the limitation of performing contact-rich manipulation tasks \cite{whitney1977force}, \cite{hogan1985impedance}. Additionally, force/motion control is crucial in robotics \cite{stemmer2006robust}. A passive compliance element (i.e. remote center compliance) in industrial robots can automate assembly tasks \cite{asada1988dynamic}, \cite{newman1999force}. However, they have to adjust according to the given applications and can only perform well on small deviations in the assembly tasks. To generalize the robotic assembly capabilities, assembly skills based on hybrid position/force trajectories, such as, linear, zigzag, spiral, sinus, and lissajous trajectories, have been developed for robots equipped with force and torque sensors \cite{siciliano2016springer}.  

\section{PROBLEM STATEMENT}\label{PROBLEM STATEMENT}

\subsection{Positioning Accuracy of Mobile Manipulators}
In the traditional magnetic tracking approach, the positioning accuracy of mobile manipulators exceed $\pm$5 mm, which does not meet the requirements of many industrial applications, such as precision assembly and tending \cite{su2018positioning}, \cite{schoettler2019deep}.    
Visual feedback is an effective method for overcoming position uncertainties during assembly tasks \cite{lee2019making}, \cite{zheng2017peg}. However, visual feedback includes errors introduced by imaging sensors, lenses, as well as intrinsic and extrinsic parameters \cite{li2019survey}, steering some researchers to focus more on execution methods than on using accurate vision sensors\cite{li2019survey}.
\subsection{Positioning Uncertainty of Flexible End Effector}
Vacuum grippers are employed to grasp horizontal or vertical flat objects using suction cups because they can avoid strong contact forces between the object and gripper \cite{jaiswal2017vacuum}.Therefore, they are frequently used in 3C production factories \cite{zhao2017accuracy}.
\begin{figure}[htbp]
\centerline{\includegraphics[width=6cm,height=4.0cm]{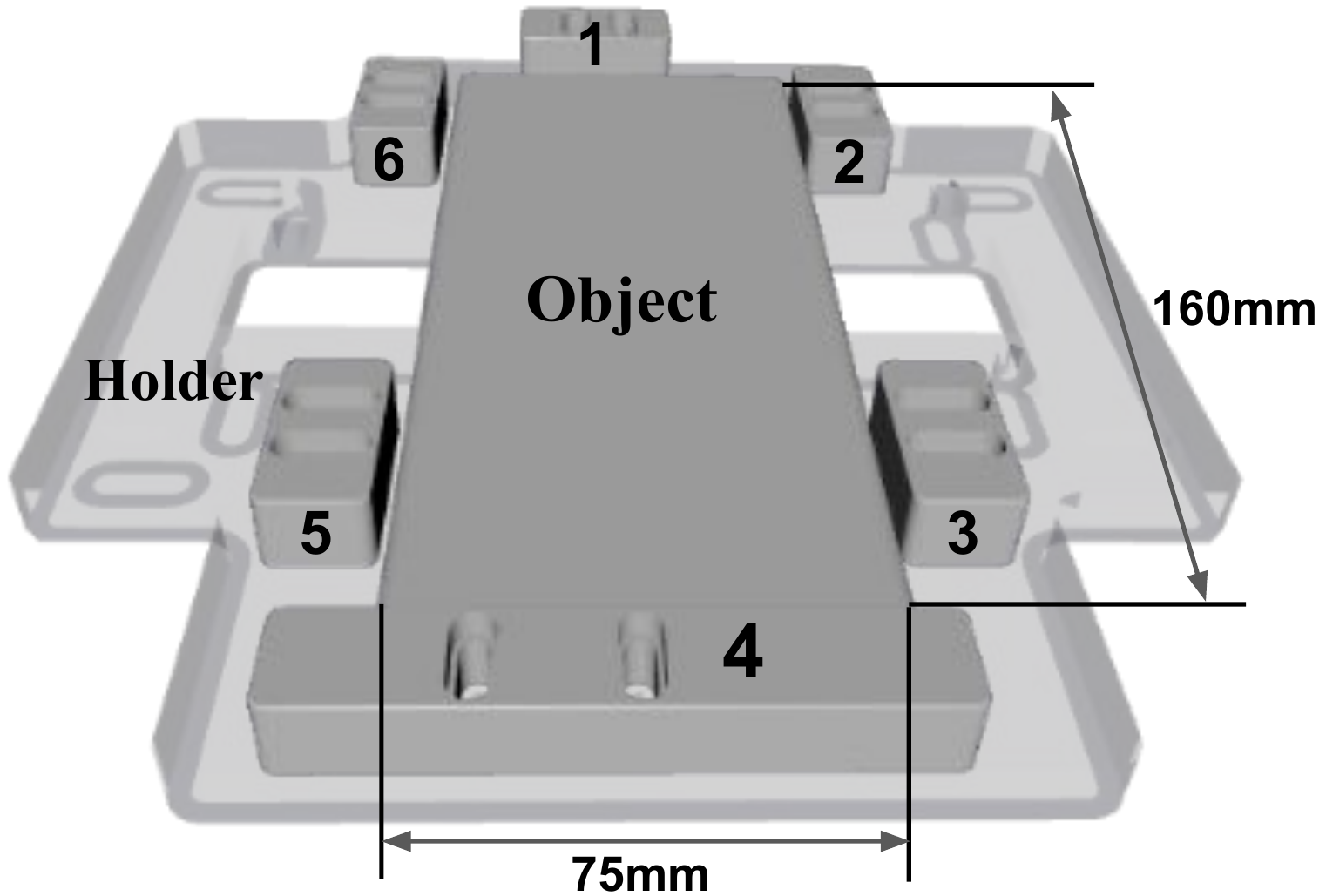}}
\caption{Assembly task illustration:  aligning two parts (object and holder) with a certain geometric feature and maintaining the alignment using a certain operation process.}
\label{fig:objectandholder}
\squeezeup
\end{figure}
\begin{figure}[htbp]
\centerline{\includegraphics[width=8.5cm,height=4cm]{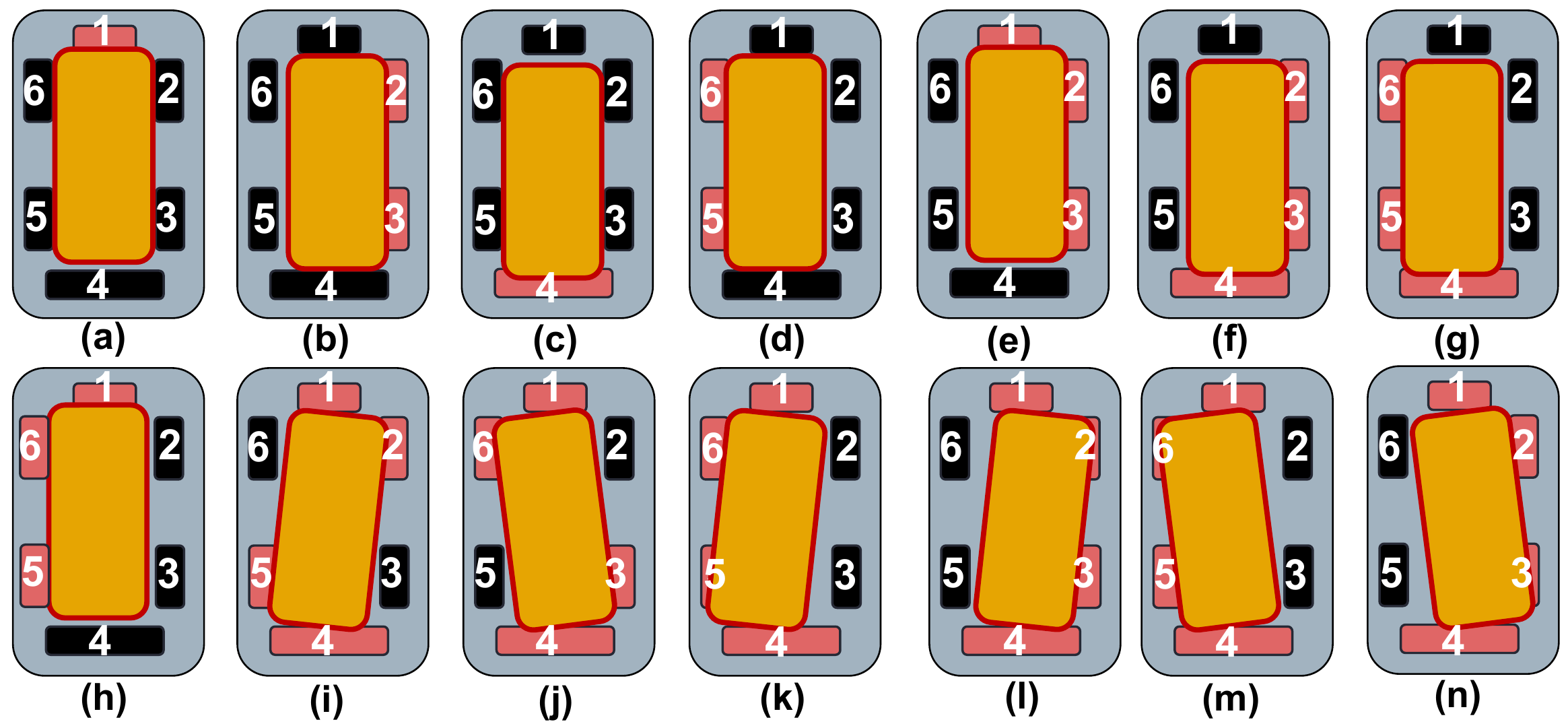}}
\caption{14 possible contact states. The pink fixtures indicate contact with the object. Due to the space limitations, some contact states were not listed.}
\label{fig:contactCases}
\squeezeup
\squeezeup
\end{figure}
Most studies consider assembly tasks as moving already-grasped objects to their goal positions by minimizing the distance between an object and its goal position \cite{luo2019reinforcement}. However, this assumption has problems owing to contact-rich tasks, particularly when using flexible vacuum grippers \cite{jaiswal2017vacuum}. 

Teach-pendant is frequently employed in factories, particularly in heavily constrained spaces, as shown in \Cref{fig:teach process}. However, avoiding strong contact forces is difficult owing to nontransparency issues (\Cref{fig:contact force curve}). Based on our experiments using a vacuum gripper (Model: Bellows Suction Cups SPB1, 1.5 Folds), a contact force of 10 N can produce a translational deformation of up to 1 mm and the respective rotational deformation is even more accentuated.
\section{CONSTRAINTS ANALYSIS AND \\SKILL DESIGN}\label{CONSTRAINTS ANALYSIS AND SKILL DESIGN}
\begin{figure}[htbp]
\centerline{\includegraphics[width=8.6cm,height=7.6cm]{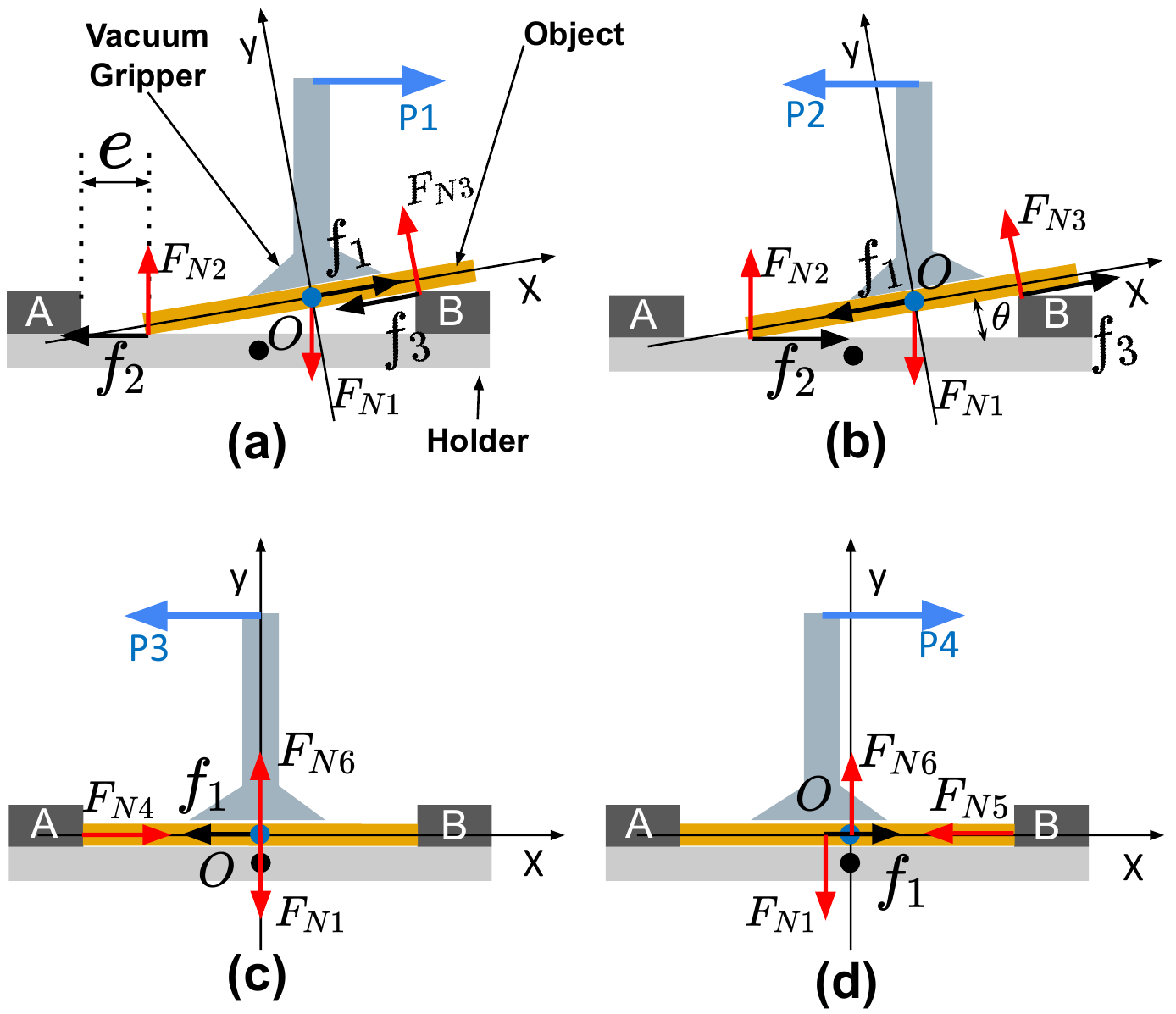}}
\caption{Illustration of a pushing action (in $X$ direction) during an assembly task. (a) The vacuum gripper pushes the object along the $X+$ direction; (b) the vacuum gripper moves to $X-$ direction, and the object can be pushed until contact occurs with the left fixture; (c) the object is pushed into the holder; and (d) the vacuum gripper continues to move along the $X+$ direction and slide on the object's surface.}
\label{fig:constraintsAnalysis}
\squeezeup
\squeezeup
\end{figure}
\subsection{Constraints Analysis}\label{Constraints Analysis}
As shown in \Cref{fig:objectandholder}, several adjustable fixtures on the holder are used to fix the object. The fixtures (numbered 1--6) are used to adjust the tension of the holder. According to our analysis, there are several contact states that can cause object and holder misalignments (\Cref{fig:contactCases}).

According to the contact states analyzed in \Cref{fig:contactCases}, fixtures may generate a strong contact force between the object and robot, causing a stuck situation. Moreover, owing to the elastic effect of the EE (e.g., vacuum gripper), the concrete contact situation is difficult to distinguish with the force/torque sensor mounted at the wrist of the robot.

Through the simulations of robot assembly manipulations with human hands, we found that the constraints provided by the fixtures can guide the object alignments with the holder geometry.
\Cref{fig:constraintsAnalysis} was used to analyze this phenomenon according to a simplified contact situation, whereas \Cref{fig:constraintsAnalysis}(a)--(d) illustrate the pushing action along the $X$ direction. 

$\boldsymbol{e}$ is the initial position error between object and holder. $\boldsymbol{P}$ is the translational movement command along the $X$ direction. Frame $O$ is attached to the center of the object, and $\theta$ is the angle between the object and holder surface. \textbf{A} and \textbf{B} represent the adjustable fixtures on the holder. $\boldsymbol{f}_1$ and $\mu_1$  represent the maximum static friction and coefficient of static friction between the vacuum gripper and object (the vacuum gripper is released), respectively; $\boldsymbol{f}_2$ and $\mu_2$ represent the static friction and coefficient of static friction between the object and holder/fixtures, respectively. $\boldsymbol{F}_{N1}$ is the force applied by the vacuum gripper on the object. Additionally, $\boldsymbol{F}_{N2}$, $\boldsymbol{F}_{N3}$, $\boldsymbol{F}_{N4}$, $\boldsymbol{F}_{N5}$ and $\boldsymbol{F}_{N6}$ are the normal forces exerted by the holder/fixtures on the object.   

We performed general force analysis on the object along the $X$ and $Y$ directions of frame $O$.  In \Cref{fig:constraintsAnalysis}(a), the object was assumed to be stationary. The force analysis performed along the $Y$ direction is represented by the following equation:
\begin{equation}
\setlength{\abovedisplayskip}{3pt}
\setlength{\belowdisplayskip}{3pt}
   \boldsymbol{F}_{N1}\cos{\theta} + \boldsymbol{F}_{N2}\cos{\theta} + \boldsymbol{F}_{N3} + \boldsymbol{f}_2\sin{\theta} = 0
  \label{force in y}
\end{equation}
\begin{figure}[htbp]
\centerline{\includegraphics[width=8.6cm,height=3.9cm]{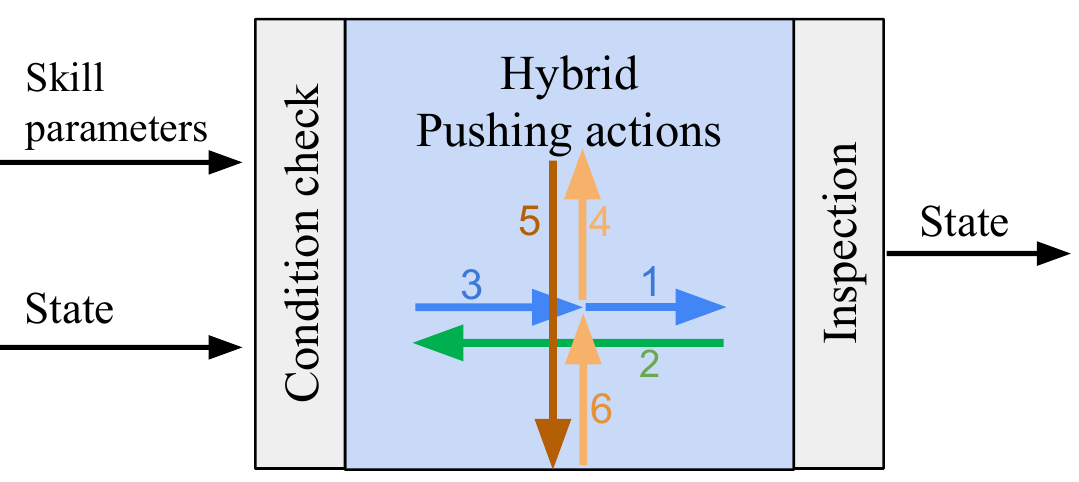}}
\caption{Pushing-based hybrid position/force assembly skill model.}
\label{fig:skillDesign}
\end{figure}
$\boldsymbol{F}_{v-o}$ represents the force between the vacuum gripper and object in $X$ direction: 
\begin{equation}
\setlength{\abovedisplayskip}{3pt}
\setlength{\belowdisplayskip}{3pt}
\begin{split}
\boldsymbol{F}_{v-o} & =  \boldsymbol{f}_1 + \boldsymbol{F}_{N1}\sin{\theta}\\
& = \mu_1\boldsymbol{F}_{N1}\cos{\theta} + \boldsymbol{F}_{N1}\sin{\theta}
  \label{v-o force in x}
\end{split}
\end{equation}
$\boldsymbol{F}_{o-h}$ represents the force between the object and holder along the $X$ direction: 
\begin{equation}
\setlength{\abovedisplayskip}{3pt}
\setlength{\belowdisplayskip}{3pt}
\begin{split}
\boldsymbol{F}_{o-h} & =  \boldsymbol{f}_2\cos{\theta} +  \boldsymbol{f}_3 \\
& = \mu_2\boldsymbol{F}_{N2}\cos{\theta} + \mu_2\boldsymbol{F}_{N3}\\
& = \mu_2(\boldsymbol{F}_{N2}\cos{\theta} + \boldsymbol{F}_{N3})
  \label{1:o-h force in x}
\end{split}
\end{equation}
Based on \Cref{force in y}, 
\begin{equation}
\setlength{\abovedisplayskip}{3pt}
\setlength{\belowdisplayskip}{3pt}
\begin{split}
\boldsymbol{F}_{o-h} & = -\mu_2\boldsymbol{F}_{N1}\cos{\theta} -  \mu_2\boldsymbol{f}_2\sin{\theta}
  \label{2:o-h force in x}
\end{split}
\end{equation}
Considering angle $\theta \approx 0$, negligible terms were removed from \Cref{v-o force in x} and (\ref{2:o-h force in x}) (when $\theta \approx 0$, $\sin\theta \approx 0$), \Cref{compare} was obtained as follows:
\begin{equation}
\setlength{\abovedisplayskip}{3pt}
\setlength{\belowdisplayskip}{3pt}
\begin{split}
&\boldsymbol{F}_{o-h} = -\mu_2\boldsymbol{F}_{N1}\cos{\theta} \\
&\boldsymbol{F}_{v-o} = \quad \mu_1\boldsymbol{F}_{N1}\cos{\theta}
  \label{compare}
\end{split}
\end{equation}
From \Cref{compare} and \Cref{fig:constraintsAnalysis}, we can infer that the coefficient of static friction determines the object's direction of motion. The same conclusion can be obtained from \Cref{fig:constraintsAnalysis}(b). The vacuum grippers are mainly made of rubber \footnote{https://www.festo.com/net/supportportal/files/216340/10783} with a high coefficient of static friction, thus $\mu_1 > \mu_2$ is an easily obtainable condition.

The following results are obtained from \Cref{fig:constraintsAnalysis}(c) and (d):
\begin{equation}
\setlength{\abovedisplayskip}{3pt}
\setlength{\belowdisplayskip}{3pt}
\begin{split}
&max(\boldsymbol{F}_{N4}) = max(\boldsymbol{F}_{N5}) \gg \boldsymbol{f}_1=  \mu_1\boldsymbol{F}_{N1}
  \label{normalforce}
\end{split}
\end{equation}
Thus, the entire operation was performed based on the following condition (\Cref{fig:constraintsAnalysis}) : 
\begin{equation}
\setlength{\abovedisplayskip}{3pt}
\setlength{\belowdisplayskip}{3pt}
\begin{split}
& max(\boldsymbol{F}_{N4}) = max(\boldsymbol{F}_{N5}) \gg  \boldsymbol{F}_{v-o} > \boldsymbol{F}_{o-h}.
  \label{rules}
\end{split}
\end{equation}
\begin{figure}[htbp]
\centerline{\includegraphics[width=8.2cm,height=3.7cm]{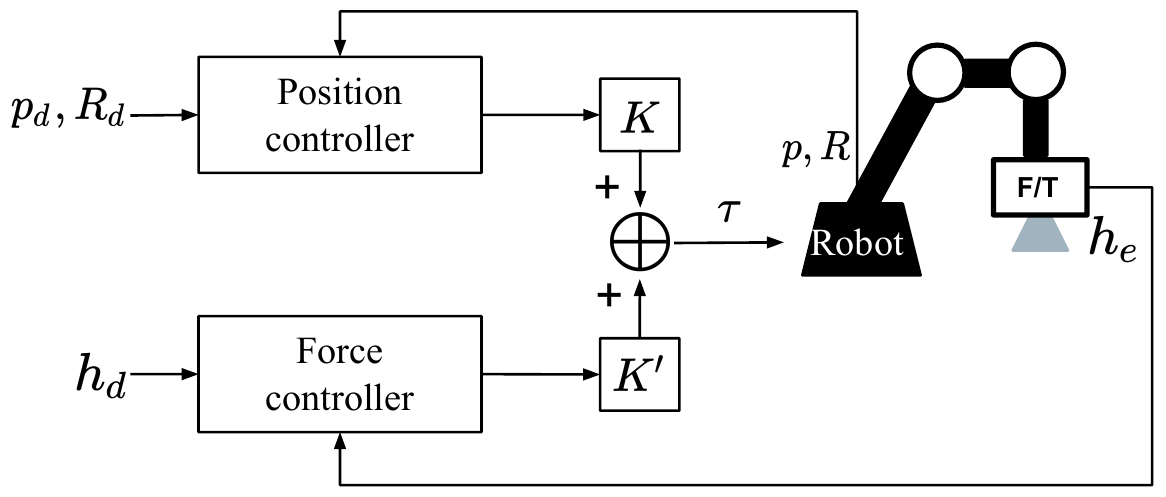}}
\caption{Hybrid position/force controller.}
\label{fig:hybridController}
\end{figure}
\begin{figure}[htbp]
\centerline{\includegraphics[width=8.0cm,height=5.1cm]{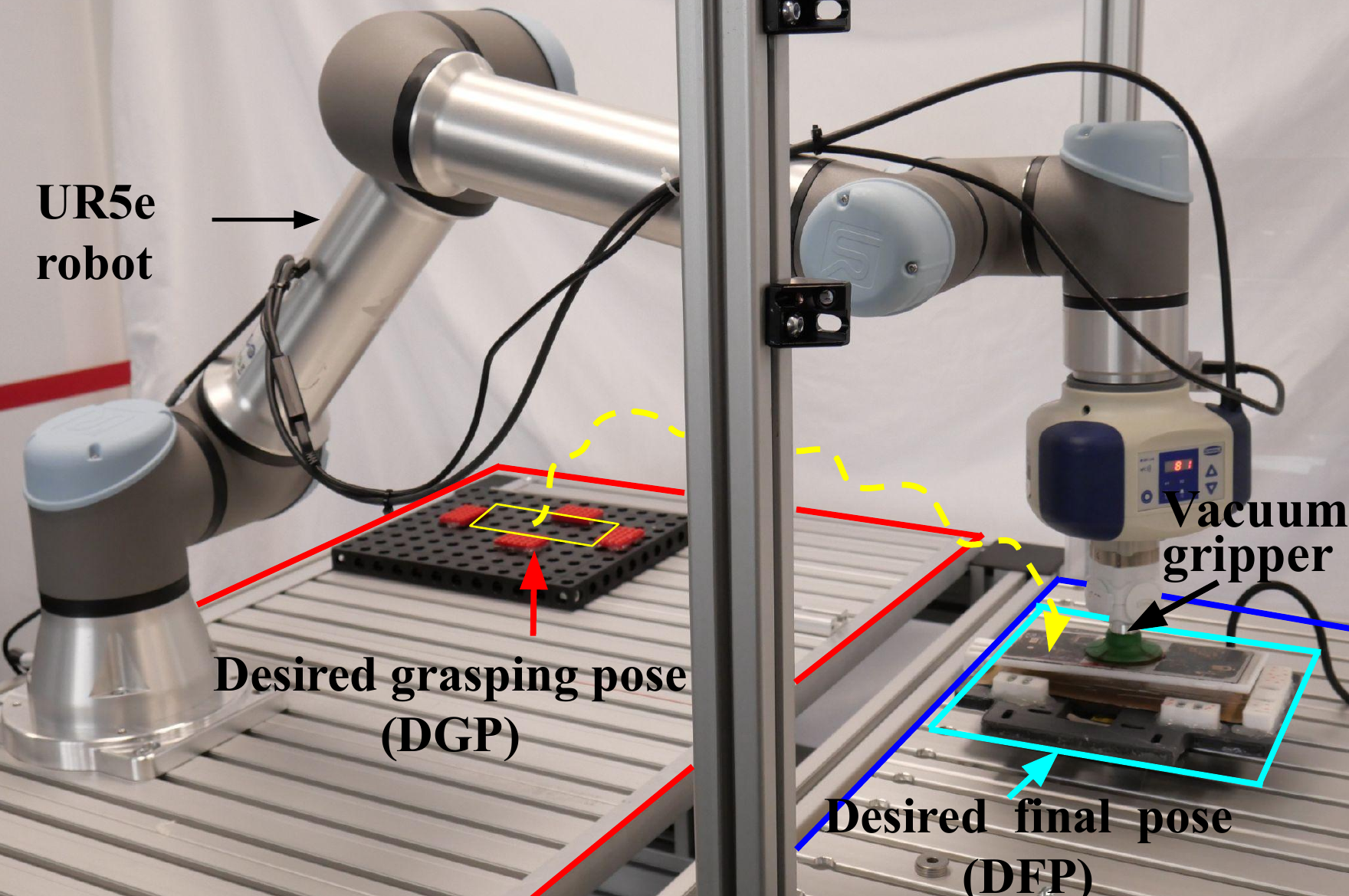}}
\caption{The robotic arm and vacuum gripper performing a contact-rich tending task.}
\label{fig:experimentSetup}
\squeezeup
\squeezeup
\end{figure}

The object can be pushed by the vacuum gripper until it is constrained by the fixtures thereby maximizing the use of environmental constraints.
Based on the aforementioned analysis, we designed a pushing-based hybrid position/force assembly skill model to execute the tending task. 
\subsection{Skill Design}\label{Skill Design}

A pushing-based hybrid position/force assembly skill was designed, as shown in \Cref{fig:skillDesign}. Once the state is confirmed, 6 linear hybrid position/force movements are executed according to the skill parameters (i.e., position and force). An inspection was performed after the execution.

We implemented a hybrid position/force controller under the EE frame for the skill execution (\Cref{fig:hybridController}), where $p_d$ and $R_d$ are command pose, $p,R$ are current pose; $h_d$ is a command force/torque, and $h_d$ is a feedback force/torque. The diagonal matrices $K$ and $K'$ were used to indicate position or force control under the EE frame. 
In this study, force control along the $Z$ direction was defined by \Cref{equ:diagonalmatrix}:
\begin{equation}
K={
\left[ \begin{array}{cccccc}
1 & 0 & 0 &0 & 0 & 0\\
0 & 1 & 0 &0 & 0 & 0\\
0 & 0 & 0 &0 & 0 & 0\\
0 & 0 & 0 &1 & 0 & 0\\
0 & 0 & 0 &0 & 1 & 0\\
0 & 0 & 0 &0 & 0 & 1
\end{array} 
\right ]},
K'=I - K
\label{equ:diagonalmatrix}
\end{equation}
\begin{figure}[htbp]
\centerline{\includegraphics[width=7cm,height=7.7cm]{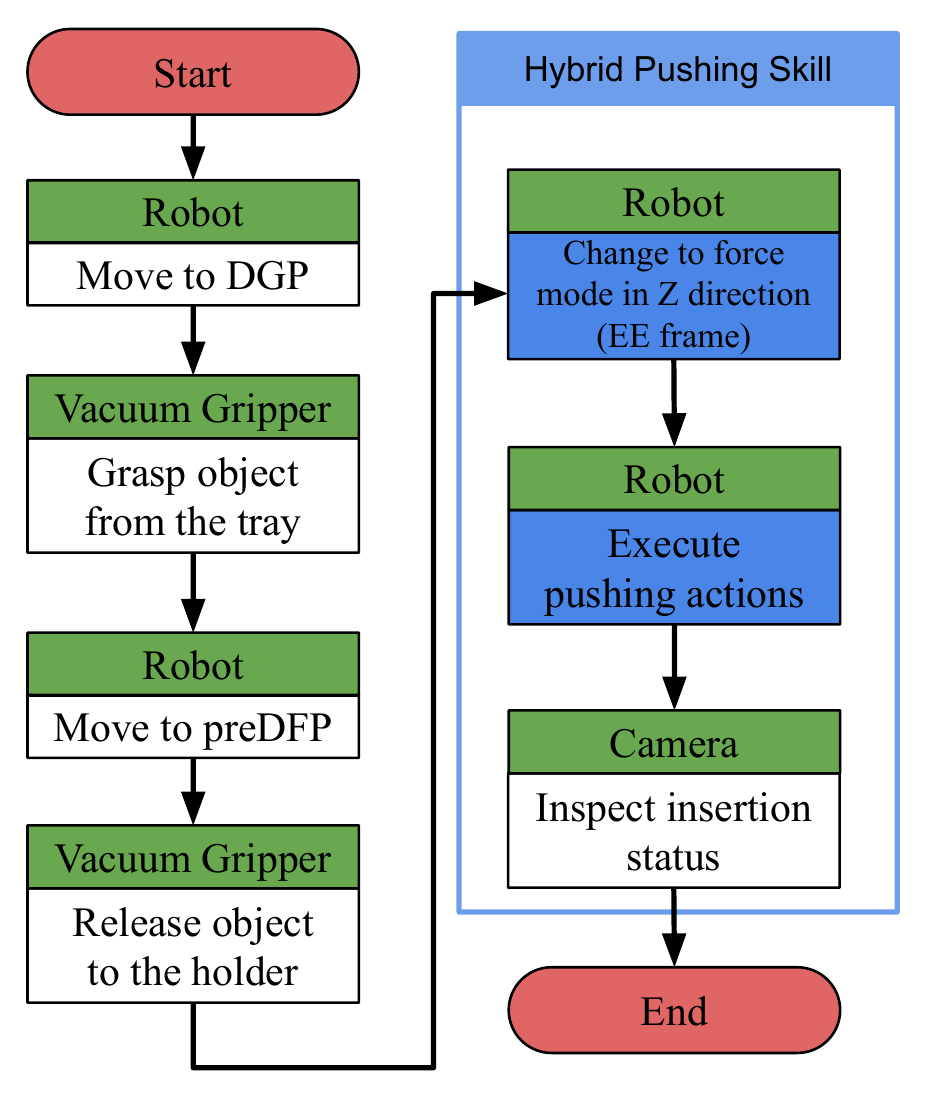}}
\caption{Workflow of a machine tending task. Before executing our hybrid pushing skill, the object was released by the vacuum gripper. DGP: desired grasping pose; DFP: desired final pose; EE: end-effector}
\label{fig:workprocess}
\end{figure}
\section{EXPERIMENTS AND CONCLUSIONS}\label{EXPERIMENTS AND CONCLUSIONS}
\subsection{Experiments}
The experiment setup can be found in \Cref{fig:experimentSetup}. A UR5e robot was used to implement the machine tending task. The UR5e features a 6-axis and 5--kg payload, a working radius of 850 mm. It is equipped with a 6-DOF force/torque sensor on its EE and uses admittance controller \cite{hogan1985impedance} to achieve operational space force control.
A workflow was designed to perform the experiment as shown in \Cref{fig:workprocess}. The skill parameters for the 6 actions (The sequence is shown in \Cref{fig:skillDesign}) are set as:
\begin{equation}\label{actions}
\setlength{\abovedisplayskip}{3pt}
\setlength{\belowdisplayskip}{3pt}
\begin{split}
1:&[+P^d_{\sigma x}, 0, +F_z]\\
2:&[-2*P^d_{\sigma x}, 0, +F_z]\\
3:&[+P^d_{\sigma x}, 0, +F_z]\\
4:&[0,+P^d_{\sigma y}, +F_z]\\
5:&[0,-2*P^d_{\sigma y}, +F_z]\\
6:&[0,+P^d_{\sigma y}, +F_z].\\
\end{split}
\end{equation}
$P^d_{\sigma x}$ and $P^d_{\sigma y}$ denote the amplitudes of the discrete actions. $P^d_{\sigma x}$ and $P^d_{\sigma y}$ is set to 8 mm which twice bigger than the error $\boldsymbol{e}$ to ensure that the environmental constraints are fully explored. $F_z$ is set to 5 N.

In this experiment, contrary to the \textbf{perfect} group, an error of $\boldsymbol{e} \in [2,4]$ mm was added along a random direction to the desired final pose to simulate the pose uncertainties in the \textbf{uncertainty} group. We evaluated our proposed method compared with the following baselines:

\noindent \textbf{Baseline 1: spiral search \cite{jasim2014position}.} A spiral search path was used to survey the entire environment surface. Here, the maximum search radius was set to 10mm.

\noindent \textbf{Baseline 2: learning-based search.}  In this study, a double deep Q-network (DQN) with proportional prioritization \cite{schaul2015prioritized} was trained as the search policy $\pi_{\theta}(s)$. This baseline comes from the previous research \cite{shi2021combining}. The 6-dimensional force-torque vector $s = [F_x, F_y, F_z, M_x, M_y, M_z]$ under the robot EE's frame $x_e$ was sent to the double DQN, and the force actions along the $X, Y, $, and $Z$ directions were output to the robot controller.

\subsection{Conclusions}

Overall, the results of 600 group robot assembly tasks were recorded, as presented in \Cref{tab:resultbaselines}.
\newcommand{\tabincell}[2]{\begin{tabular}{@{}#1@{}}#2\end{tabular}}  
\begin{table}[t!]
	\caption{Comparison of the success rates for different baselines}
	\label{tab:resultbaselines}
	\centering
	\begin{tabular}{lcc}
		\toprule
		\textbf{\textbf{Baselines}} & {\textbf{Perfect}} & {\textbf{Uncertainty}}\\
		\midrule
		Baseline 1 & 69/100  & 47/100 \\
		Baseline 2 & 95/100  & 91/100 \\
	    \textbf{Our method}	& \textbf{100/100} & \textbf{100/100} \\
		\bottomrule
	\end{tabular}
\squeezeup
\squeezeup
\end{table}
Baseline 1 (spiral search) always ended when stopped by one or two fixtures and thus insertion failed, moreover, it always generates sufficiently strong contact force between the objects and the holder. Baseline 2 (learning-based search) offered a high success rate. However, because Baseline 2 uses a neural network to estimate the state, it may sometimes lead to unexpected actions and thus failed. Our method achieves the best success rate in this comparison experiment.

To verify the generalization of our method, we also tested it on two other type holders using our method and obtained a 100\% success rate (100/100 trials). Due to space limitation, we do not describe the details in this paper. The test scenes can be seen in the video submitted with the paper.

In future work, the aforementioned results will be combined with other learning-based pushing methods \cite{cong2020self} (our previous work) to improve the efficiency and safety of the tending task.

\section*{Acknowledgment}
We would like to thank Jingjing Sun for her contributions as a robot teaching process volunteer. We also thank mechanical engineer Ningxin Lu for building the hardware mockup system.

\balance

\bibliographystyle{IEEETran}
\bibliography{robio2021} 

\end{document}